# Missing Data Prediction and Classification: The Use of Auto-Associative Neural Networks and Optimization Algorithms

Collins Leke*, Bhekisipho Twala, and T. Marwala

**Abstract**

This paper presents methods which are aimed at finding approximations to missing data in a dataset by using optimization algorithms to optimize the network parameters after which prediction and classification tasks can be performed. The optimization methods that are considered are genetic algorithm (GA), simulated annealing (SA), particle swarm optimization (PSO), random forest (RF) and negative selection (NS) and these methods are individually used in combination with auto-associative neural networks (AANN) for missing data estimation and the results obtained are compared. The methods suggested use the optimization algorithms to minimize an error function derived from training the auto-associative neural network during which the interrelationships between the inputs and the outputs are obtained and stored in the weights connecting the different layers of the network. The error function is expressed as the square of the difference between the actual observations and predicted values from an auto-associative neural network. In the event of missing data, all the values of the actual observations are not known hence, the error function is decomposed to depend on the known and unknown variable values. Multi-layer perceptron (MLP) neural network is employed to train the neural networks using the scaled conjugate gradient (SCG) method. Prediction accuracy is determined by mean squared error (MSE), root mean squared error (RMSE), mean absolute error (MAE), and correlation coefficient (r) computations. Accuracy in classification is obtained by plotting ROC curves and calculating the areas under these. Analysis of results depicts that the approach using RF with AANN produces the most accurate predictions and classifications while on the other end of the scale is the approach which entails using NS with AANN.

## Introduction

Inferences and conclusions drawn from datasets with all required data entries are likely to be more reliable and useful than those which emerge from incomplete datasets. With this in mind, there are several ways in which the problem of missing data may occur in a dataset. Most significant ways entail data entry errors or failures at sensors which are meant to record the data. In this paper, the aim is to predict or classify a specific variable with all other variables present and complete. In the event of there being missing values in variables meant to be complete, missing data imputation techniques such as mean substitution or regression models are used to complete the data prior to the prediction or classification tasks. In a dataset with several input variables and a continuous prediction variable, the problem is how to determine the missing values for the prediction variable having some prior knowledge of the relationships between the input variables and the prediction variable. Are there any mechanisms to accurately predict the missing prediction variable values based on the observed relationships between variables?
Furthermore, in a dataset with several input variables and one discrete output/prediction variable represented ideally numerically by 0 or 1 in a two class case for easy analysis, given that all other data entries are provided and the only missing data is that of the discrete prediction variable, is there a means of determining this value correctly, thereby assigning an unseen data record into the correct class (0 or 1)? Moreover, we investigate and evaluate the capabilities of these algorithms in solving these kinds of problems on the respective datasets. With these known from the experiments performed in this paper, a comparative statistical analysis of the results obtained by these algorithms on the same dataset is performed to identify which algorithm yields the best results on the given dataset.

## Background

### A. Network design

#### 1. Auto-Associative Neural Network

Neural Networks are information processing models that simulate the manner in which biological nervous systems process information [Mistry et al. 2008]. There are four main constituents of any neural network and they include the processing units, activation functions, weighted interconnections and the activation rules [Mistry et al. 2008]. Auto-associative neural networks (AANN) are network models in which the network is trained to recall the inputs as the outputs (Lu and Hsu, 2002), thus guaranteeing the networks are able to predict the inputs as outputs whenever new inputs are presented. These networks have been used in a variety of applications [Atalla and Inman, 1998; Froloy et al. 1995; Smauoi and Al-Yakoob, 2003; Hines et al. 1998; Marwala 2001; Marwala and Chakraverty, 2006; and Brian et al. 2006]. An auto-associative network encoder also referred to as an auto-encoder consists of an input and output layer with the number of inputs being equal to the number of outputs, hence the name auto-associative [Lu and Hsu, 2002]. In addition to these two layers, there also is a narrow hidden layer. It is necessary that the hidden layer uses a lower dimension, so as to enforce the encoding and decoding processes [Mistry et al. 2008]. This hidden layer attempts to reconstruct the inputs to match the outputs by minimizing the error between the inputs and outputs with the introduction of new inputs. The narrow hidden layer forces the network to reduce any redundancies that may occur in the data whilst allowing the network to detect non-redundant data [Baek and Cho, 2003]. Figure 1 depicts the framework of an AANN. For the purpose of this work, one hidden layer is used as it has been proven that a single hidden layer network is capable of approximating any continuous multivariate function to any suitable degree of accuracy [Tim et al., 2004]. The number of nodes in the hidden layer is determined by the ability of the network to approximate the error function.

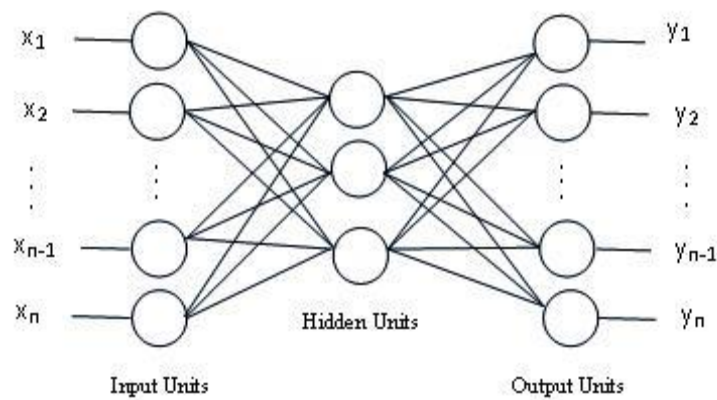

Fig. 1: Auto-Associative Neural Network Structure

#### 2. Multi-Layer Perceptron

These networks consist of multiple layers of computational units usually interconnected in a feed-forward manner [Mistry et al. 2008]. These networks have been used in several papers due to their stability and ease of use [Brian et al., 2006; Abdella and Marwala, 2005; and Tim et al., 2004]. They require an expected target in order to be trained and are therefore referred to as supervised learning networks due to this trait [Paul and Mitra, 2002]. A fully connected two layered MLP architecture is used in this paper. A NETLAB toolbox (Nabney, 2003) that runs in MATLAB is used to implement the network architecture. Figure 1 depicts the architecture of the MLP which is the platform on which the AANN is designed and trained. MLP networks learn how to transform an input data point into a target data point hence they are widely used for pattern classification [Brian et al., 2006]. A variety of learning techniques are used in MLP networks, the most frequently applied being back-propagation. In back-propagation, the obtained output values at the output units are compared with the actual output values to calculate the value of a predefined error function. The error is then fed-back into the network and is used by the algorithm to adjust the weights of each connection accordingly, so as to minimize the value of the error function [Brian et al., 2006]. This process, referred to as training the network, is done iteratively as the error function value tends to converge to a number small enough that it no longer changes with changing iterations. The MLP network can be described as per equation (1) [Ming-Hau, 2010]:

$$y_k = f_{outer}\left(\sum_{j=1}^{M} w_{kj}^{(2)} f_{inner}\left(\sum_{i=1}^{d} w_{ji}^{(1)} + w_{j0}^{(1)}\right) + w_{k0}^{(2)}\right) \qquad (1)$$

Here $y_k$ represents the $k^{th}$ output, $f_{outer}$ represents the output layer transfer function, $f_{inner}$ represents the input layer transfer function, $w$ represents the weights and biases, with the superscripts 1 and 2 representing the first and second layers respectively.

### B. Algorithms

#### 1. Genetic Algorithms

Genetic Algorithms (GAs) form part of a subclass of computational evolutionary algorithms in which the elements of the solution space have binary string representations. They can be defined as algorithms tasked with finding approximate solutions to complex problems through the application of evolutionary biology principles [Abdella and Marwala, 2005]. They make use of techniques derived from biological processes like inheritance, mutation, natural selection, and recombination [Abdella and Marwala, 2005]. Genetic algorithms are inspired by Darwin's theory about evolution. Solution to a problem solved by genetic algorithms is evolved. GAs have been proven to be successful in optimization problems like wire routing, scheduling, adaptive control, cognitive modelling, travelling salesman problems, optimal control problems and database query optimization problems [Abdella and Marwala, 2005]. GA begins with a **set of solutions** (represented by **chromosomes**) called **population**. Solutions from one population are taken and used to form a new population. This is motivated by a wish, that the new population will be better than the old one. Solutions which are selected to form new solutions (**offspring**) are selected according to their fitness - the more suitable they are the more chances they have to reproduce. This procedure is repeated until some condition (for example number of populations or improvement of the best solution) is satisfied.

Genetic Algorithm therefore does not only demarcate itself in terms of its approach from conventional optimization techniques, but also offers another technique for situations where conventional methods are inefficient and inadequate. There are four ways in which genetic algorithms are different from conventional optimization methods and these are [Goldberg, 1989; Michalewicz, 1996; and Forrest, 1996]:

- GA encodes solutions using a binary string representation, and this is the encoding which the algorithm works with. All of the individuals in the population are an encoding of a possible solution to the optimization problem being analysed;
- GA works concurrently with a population of solutions and not just one candidate solution;
- GA uses only a fitness function to perform the evaluation of possible solutions within the population and not any other information;
- GA incorporates an element of randomness to the search procedure, not solely dependent on deterministic rules.

The main steps involved in the genetic algorithm implementation for solving a particular problem are as follows [Goldberg, 1989; and Michalewicz, 1996]:

- Generate an initial random population of potential candidate solutions with an adequate problem specific representation;
- Generate an evaluation function and calculate the fitness value for each of the elements of the population;
- Create offspring by using three genetic operators (reproduction, crossover and mutation) which are determined such that they alter the composition of each offspring;
- Determine various parameters that the genetic algorithm uses (population size, probabilities of applying genetic operators, crossover probability, and mutation probability and so on).
- Evaluate the new solution (offsprings) and calculate the fitness of each solution and;
- If optimum solution is achieved, stop and return, otherwise create new offspring as in step 3 and evaluate their fitness.

A pseudo-code for the genetic algorithm which illustrates a detailed description of how the genetic algorithm works can be found in Michalewicz (1996).

## 2. Simulated Annealing

Simulated annealing is a probabilistic algorithm tasked with finding the global minimum of a function that may have several local minima [Dimitris and John, 1993]. It is also referred to as a random search algorithm that makes use of an analogy similar to the manner in which a metal loses heat and produces a low energy crystalline structure from a high heat and energy state, the annealing schedule, and the manner in which a discrete optimization problem is minimized [Busetti, 2012]. The atoms of the metal initially have high temperatures and this allows them to move freely and restructure themselves at will. A decrease in the temperature results in a decrease in the energy levels of these atoms to a state in which a minimum energy level is attained. The algorithm starts with the system having a very high temperature in a similar manner whereby the input values are allowed to take on a great range of variation in the values. The temperature is allowed to fall as the algorithm progresses and this enforces a limit on the degree to which the variable values are allowed to be dissimilar. This aspect of the algorithm often leads to the finding of a better solution. As far as there is a guarantee that there is a reduction in the temperature, there will be changes created in the algorithm inputs, producing successively better solutions and giving rise to an optimal set of values when the temperature is at its lowest. If the cooling process is sufficiently slow, the final configuration of the metal being cooled is such that it results in a solid with such superior structural integrity that it is free of all crystalline defects [Busetti, 2012].

Simulated Annealing has an advantage over other optimization techniques in that it is able to avoid being stuck in a local minimum [Busetti, 2012]. Each iteration of the simulated annealing algorithm applied to a discrete optimization problem guarantees that the objective function or fitness function generates values for two solutions which are compared to each other. Better solutions are always accepted, while a fraction of the not so good solutions are accepted with a probability in order to prevent the possibility of the algorithm being stuck in local minima in the search of global minima. The main disadvantage of this algorithm as commonly observed with stochastic local search algorithms is that defining some of the key parameters such as the initial temperature and cooling rate are subjective and must be done from an empirical basis.

The SA algorithm is a technique which is capable of [Ingber, 1993]:

- Processing evaluation functions which possess arbitrary degrees of non-linearities, discontinuities and which are of a stochastic nature;
- Processing quite arbitrary boundary constraints which are mandatory for the evaluation functions;
- Being implemented with relative ease, and with minimal coding when compared with other non-linear optimization techniques like genetic algorithms and;
- Statistically guaranteeing the finding of a global optimal solution.

For this algorithm to work efficiently and always guarantee the return of a global optimal solution and not a local optimal solution some key elements are required and they are [Busetti, 2012]:

- An adequate representation of the candidate solutions;
- A means of generating changes to these candidate solutions, and they should be random;
- An evaluation mechanism for the solutions using a fitness function and;
- An annealing schedule which involves choosing an initial temperature and rules for lowering it as the search process progresses.

In spite of the positive aspects this algorithm presents, there is a weakness of the algorithm in terms of the obvious trade-off between the quality solutions obtained and the time it takes to compute them. More of the weaknesses of the algorithm could be found in Busetti (2012). Attaining an optimal solution by use of the simulated annealing algorithm could take an infinite amount of time, and this could be attributed to the process of carefully controlling the rate at which the temperature cools down. To overcome this problem, Fast annealing (FA), very fast simulated re-annealing (VFSR) or adaptive simulated annealing (ASA) which are progressively faster could be used [Busetti, 2012].

### 3. Particle Swarm Optimization

Particle Swarm Optimization is a computational algorithm that optimizes a problem by iteratively making an attempt to improve a candidate solution taking into account a specific measure of the quality of the solution [Birge, 2003]. The particle swarm optimization algorithm is based on two fundamental principles being social science and computer science [Del Valle et al., 2008]. Furthermore, PSO makes use of the swarm intelligence principle which is the characteristic of an environment in which the overall behaviours of primitive individuals that interact locally with one another and with their environment create coherent general functional patterns of behaviour. The optimization is achieved by having an initial population of candidate solutions, moving these around in a solution-space according to a simple mathematical formula, in this case the error/fitness function to be minimized, taking into account the particle's position and velocity [Poli et al., 2007]. This algorithm can be thought of as a meta-heuristic as it makes few or no assumptions at all about the problem being optimized and can search very large spaces of candidate solutions [Poli et al., 2007]. In PSO literature, the term particles refers to a member of the population of candidate solutions which have no mass and volume, and are influenced by their velocities and accelerations towards a better behaviour.

The approach used by the particle swarm optimization algorithm in solving a problem emerges from the interactions of individual particles in a population as well as the particles themselves, in a population wide phenomenon. These populations of particles form some kind of interconnected topology which is often referred to as a social network. The interconnected structure is ideally made up of two-way lines linking immediately adjacent particles, in such a way that if j is in i's neighbourhood, then i will also be in j's neighbourhood. There is communication between each of the particles and it is affected by the best point found by any member of its topological neighbourhood. This is the vector $\vec{p}_i$ for that best neighbour, which will be denoted by $\vec{p}_g$. There have been a wide variety of population structures being applied in the implementation of this algorithm with some of the most commonly used ones being the ring topology or the wheel topology.

In the PSO algorithm process, the velocity of each particle is iteratively adjusted so that the particle stochastically oscillates around the locations of $\vec{p}_i$ and $\vec{p}_g$. The process for implementing the PSO algorithm is presented below [Poli et al., 2007]:

- Initialize a population of particles with random positions and velocities on $D$ dimensions in the search space:
- **Loop:**
- For each particle, evaluate the desired optimization fitness function in $D$ variables:
- Compare the different particle's fitness evaluation with its $pbest_i$:
    - If current value is better than $pbest_i$, then set $pbest_i$ equal to the current value and $\vec{p}_i$ equal to the current location $\vec{x}_i$ in $D$-dimensional space:
- Identify the particle in the neighbourhood with the best success so far, and assign its index to the variable $g$:
- Change the velocity and position of the particle according to the following equation:

$$\begin{cases} \vec{v}_i \leftarrow \vec{v}_i + \vec{U}(0,\phi_1) \otimes (\vec{p}_i - \vec{x}_i) + \vec{U}(0,\phi_2) \otimes (\vec{p}_g - \vec{x}_i), \\ \vec{x}_i \leftarrow \vec{x}_i + \vec{v}_i \end{cases}$$

- If a criterion is met, usually a sufficiently good fitness function value or a maximum number of iterations, exit loop:
- **end loop**

## 4. Random Forest

Random Forest (RF) is a method for classification and regression which was introduced by Breiman and Cutler and it is referred to as a learning ensemble consisting of a bagging of unpruned decision trees with a randomized selection of features at each split [Palmer et al., 2007]. Furthermore, as per Liaw and Wiener (2002), it could be referred to as an ensemble classifier that consists of many decision trees and outputs a prediction of a continuous variable which is obtained as the average of the prediction of all trees. This algorithm is widely used due to a relatively high level of accuracy in its predictions [Pantanowitz and Marwala, 2008]. A decision tree is one of the most popular learning methods commonly used for data exploration. It is a tree with nodes which represent information corresponding to attributes in the input vectors [Pantanowitz and Marwala, 2008]. The trees used in this algorithm are referred to as Classification and Regression Trees (CART). The prediction of missing data values is achieved by the algorithm creating a model which comprises of a number of CART trees as set by the programmer, training this model using the training dataset, and then testing the models prediction and classification capabilities using the test set. The predictions and classifications are attained by finding the averages of all the predictions of the individual CART trees created which best minimize the error function.

It has been observed in recent studies that the Random Forest algorithm presents characteristics which make it very appealing for different areas of application. These features include relatively high levels of accuracy in predictions, built-in feature selection capabilities, and a mechanism for evaluating the influence of each feature or variable to the algorithm [Palmer et al., 2007]. It also provides a paradigm to carry out data visualization for high dimensional data; clustering, anomaly, outlier, and error detection; and explicit missing data imputation. Classification and Regression Trees (CART) are known to be fast and possess excellent accuracy levels [Pantanowitz and Marwala, 2008]. Random Forest regression theoretical background could be seen in the papers of Svetnik et al. (2003), Biau (2012), Breiman (2001), and Breiman and Cutler (Breiman and Cutler, 2012). This technique is based on the creation of an ensemble of decision trees which are used to carry out predictions of continuous variables by averaging the predictions of all trees, and a classification is obtained by use of a weighted or unweighted majority voting mechanism. In order to grow this ensemble of decision trees, random vectors are often generated that govern the growth of each tree in the ensemble [Breiman, 2001]. An early example is bagging, a scenario in which to grow each tree, a random selection is made from the data samples in the training set of data without any replacement [Biau et al., 2008; Breiman, 1996]. Yet another example is the random split selection approach in which at every node, a split is chosen at random from among the K-best splits identified [Dietterich, 2000]. There is another approach which entails generating new training set samples by randomly selecting outputs from the initial training set of data.

Breiman (2001) suggested the random forest algorithm which adds an extra layer of randomness to the bagging process. The random forest algorithm also changes the way in which the classification and regression trees are created in addition to using a separate bootstrap sample of the data to create the trees. The procedure with standard trees is such that all of the nodes are split by making use of the best possible split from the entire list of variables and this radically speeds up the tree growing process. In a random forest, the procedure however is such that every node makes use of the best possible split from a subset of predictors selected at random at the node to carry out the node splitting procedure. The best splitter from the eligible random subset used to split the node might either be the best overall, or just a fairly good splitter, or may not be of any help at all. If the splitter is not very helpful, the outcome from the split is two nodes which are essentially similar. In addition to the previously mentioned benefits of the random forest algorithm, it is very easy to implement based on there being only two important control variables being the number of variables in the random subset to use for splitting at the nodes and the number of trees in the forest, and it has been observed that the algorithm is not influenced to a great extent by the values of these [Liaw and Wiener, 2002].

## 5. Negative Selection

The negative selection algorithm is based on the self/non-self discrimination principle or, in other words, the ability of the body to tell what the difference is between a self and a non-self-substance [Delahunty and Callaghan, 2004]. As stated above, the screening procedure that occurs within the body in the thymus gland is the basis on which this algorithm is designed. T-cells which are produced in the bone marrow are moved to the thymus where they undergo a maturation process in which the cells that generate responses to harmless substances are expelled and only the cells that do not adhere to the harmless substances remain. This procedure guarantees that the T-cells that remain are capable of recognizing only foreign or non-self-cells. These mature cells are then released into the body and eliminate any substances they can adhere to. Any substance that adheres to these mature cells and gets deleted is referred to as a non-self-substance.

The negative selection algorithm uses a process which is analogous to that mentioned above so as to generate its negative detectors. A set of detectors are initially generated at random. All of the detectors then go through the maturation phase in which if any of these are observed to be similar to any of the components of the set of self substances, they are eliminated and substituted by a different detector which is generated randomly. This guarantees that the size of the set of detectors is unchanged throughout the maturation process. These newly generated detectors which are meant to replace the eliminated ones are also checked against the set of self-substances. The result from this entire process is mature detectors which are those that detect potentially non-self-substances only. The procedure is done repeatedly until the result is a set of mature detectors that only detect non-self-substances.

After the creation of these mature detectors, they are presented new set of data elements. All of the mature detectors are used to try find matches against each and every component of the test set of data. Every time a match is found, it is assumed that a non-self-substance has been identified. This attribute is just one of the many advantages of the negative selection algorithm as it ensures the identification of unknown substances to the system.

## Methodology

### A. Experimental Setup

The NETLAB toolbox [Nabney 2003] is used to create the auto-associative neural network. This toolbox has a 2-layer MLP network, which according to literature review (Bishop 1995) is capable of modelling any complex relationship, such as the financial model. The network implemented consist of an input layer, representing different inputs and prediction variable, mapped to an output layer representing the same characteristics as the input layer via the hidden layer. The network is thus trained to recall itself. One of the input nodes in the input layer represents the prediction variable, which is ultimately represented by one of the output nodes as well. The neural network equation can be written mathematically as in equation (2).

$$\{y\} = f(\{x\}, \{w\}) \qquad (2)$$

Since the network is trained to recall the inputs, the output vector $\{y\}$ (predicted inputs) obtained will be approximately equal to the input vector $\{x\}$ (actual input values). The weight vector is represented by $\{w\}$. An error, however, exists between the input vector $\{x\}$ and the output vector $\{y\}$ due to the fact that they would not always be equal and this error can be expressed as the difference between the input and output vector. This error is formulated as:

$$e = \{x\} - \{y\} \qquad (3)$$

Substituting for $\{y\}$ from equation (2) into equation (3) we get:

$$e = \{x\} - f(\{x\}, \{w\}) \qquad (4)$$

In this work, a minimum and non-negative error will be used. This can be obtained by squaring the error function in equation (4) to obtain:

$$e = (\{x\} - f(\{x\}, \{w\}))^2 \qquad (5)$$

In the case of missing data and in an attempt to predict this missing data, one of the inputs in the input vector $\{x\}$ was assumed as an unknown input, while the other input properties were considered as the known inputs. When the input vector $\{x\}$ has unknown elements, the input vector set can be categorized into $\{x\}$ known represented by $\{x_k\}$ and $\{x\}$ unknown represented by $\{x_u\}$. Rewriting (4) in terms of $\{x_k\}$ and $\{x_u\}$, we obtain:

$$e = \left( \begin{pmatrix} x_u \\ x_k \end{pmatrix} - f\left( \begin{pmatrix} x_u \\ x_k \end{pmatrix}, \{w\} \right) \right)^2 \qquad (6)$$

Here $\{x_u\}$ represents the unknown input parameter, $\{x_k\}$ represents the known input parameters, $\{w\}$ represents the weight vector that maps the auto-encoder network input vector $\{x\}$ to the same input vector $\{x\}$.

An estimated value for the prediction variable will then be obtained by choosing the value which is closest to the actual test set value, and which best minimizes equation (6) using the five algorithms. GA, however, always finds the optimal maximum value. To cater for this so that it finds the optimal minimum value instead, the negative of equation (6) is used. SA, PSO, RF, and NS do find the minimum value and as such require no alteration to the minimization of the error function given in equation (6). The error function to be minimized by GA is thus:

$$e = -\left( \begin{pmatrix} x_u \\ x_k \end{pmatrix} - f\left( \begin{pmatrix} x_u \\ x_k \end{pmatrix}, \{w\} \right) \right)^2 \qquad (7)$$

The algorithms are used to solve two problems being prediction and classification problems. Prediction or forecast is a representation of the manner in which things can occur in the future, almost but not always dependent on past experiences and occurrences. Classification refers to categorization which is a process whereby ideas and objects are recognized, differentiated, understood, and grouped.

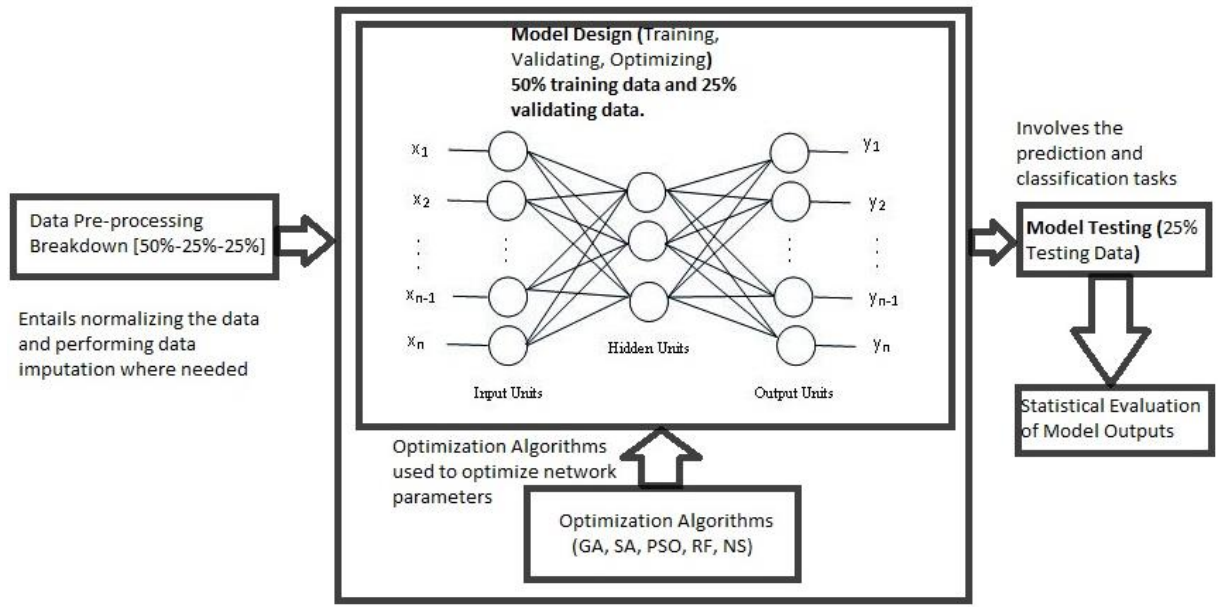

Figure 2: Diagrammatic Representation of Experimental Setup

B. Problem Statement

Although humans provide an immediate means of performing prediction and classification tasks, there are sometimes doubts with regards to the quality and accuracy of the predictions and classifications they produce. This brings forth the need for a more efficient and accurate means of performing these tasks. The use of historical data to create models that yield accurate outputs, and which can then be used on more recent and current data to perform these tasks is a growing area of research. There are several papers which have made use of AANNs to do predictions and classifications and have reported high accuracy levels. In this study, we use AANNs with five optimization algorithms (GA, SA, PSO, RF, and NS) to extract the nonlinear patterns from the input data, and use them to predict and classify specific prediction or classification variables from four separate datasets.

C. Data Pre-Processing

Three distinct datasets which vary in the number of data points and attributes are used in this study to solve prediction and classification problems. The Heart Disease dataset has 14 variables, 13 of these being independent variables, and the fourteenth being dependent on these and represents the health status of an individual. This variable has two classes (categories) only. There are categorical variables in the classification datasets which are the German Credit and Heart Disease datasets. These datasets however also have nominal, binary, and numeric variables. The German Credit dataset has a total of 25 variables with one of these representing the credit status of an individual. The Forest Fire dataset has 13 variables in total with the thirteenth, the area, being the prediction variable dependent on the other 12 variables. There are no categorical variables in this dataset. There also is the presence of outliers in the datasets, which are handled by normalizing the data to being within a given range, in this case between 0 and 1. The data is normalized using equation (8):

$$x_{scaled} = \frac{x - x_{min}}{x_{max} - x_{min}} \tag{8}$$

Where x is the value of a specific variable being considered, $x_{min}$ is the smallest value for a specific variable, and $x_{max}$ is the biggest value for the same variable. This equation ensures that all the values are in the range [0; 1]. The usual procedure for division of the data into three parts is followed, to assess the generalization ability of the network model. The latest 25% of the data is reserved for testing in all four datasets, and the remaining portion is divided into a training and validation dataset. The full set is split as follows for all four datasets used: 50% for training, 25% for validation and 25% for testing.

### D. Experiment 1: German Credit Data

German Credit data from the UCI data repository [UCI, 2012] is used to train the network, and test the classification capabilities of the methods. The variables differ in range and data types but normalization of the values results in uniformity of these. There are no variable names used in this dataset for confidentiality reasons and as such the variables are represented by A1 to A25, representing all 25 attributes. A Multi-layer perceptron with 25 inputs, 25 outputs and 12 hidden nodes in one hidden layer was trained on the data. The number of hidden nodes in the network is obtained through an iterative process which tests the network for a number of hidden nodes from 2 to n - 1, in this case 25 - 1 = 24. That which results in the lowest network error is chosen as the optimal number of hidden nodes, in this case 12, and it is problem specific. A total of 1000 data points for each of the 25 variables were provided with 500 of each of these, accounting for 50% of the dataset used to train the network. The classification variable, being the Credit Status (A25), was assumed to have all 250 (25%) of its test data entries missing at random and the aim was to accurately predict and classify these from having prior knowledge of how the other variables affect the outcome. To measure the level of accuracy with which these algorithms are able to predict the expected outputs and classify individuals correctly, Receiver Operating Characteristic curves are plot and the areas under these are calculated.

### E. Experiment 2: Forest Fire Data

An MLP neural network with 13 input units, seven hidden units in one hidden layer, and 13 output units were trained on the training set of the Forest Fire dataset from the UCI data repository [UCI, 2012]. A total of 259 (50% of dataset) training input patterns were provided for the network architecture. The elements from the test set (129 in total) for the prediction variable, the Area affected by fire (A13) in this case were removed and assumed missing and these were to be approximated by the algorithms. To assess the effectiveness of the algorithms in approximating the missing values, the mean squared error, root mean squared error, and mean absolute error were calculated for each missing value.

### F. Experiment 3: Heart Disease Data

Heart Disease data from the UCI data repository [UCI, 2012] is used to train the network, and test the classification capabilities of the algorithms. The variables differ in range and data types but normalization of the values results in uniformity of these. The variables are represented by A1 to A14, representing all 14 input variables. A Multi-layer perceptron with 14 inputs, 14 outputs and eight hidden nodes in one hidden layer was trained on the data. The number of hidden nodes in the network is obtained through an iterative process which tests the network for different numbers of hidden nodes from 2 to n - 1, in this case 14 - 1 = 13. That which results in the lowest network error is chosen as the optimal number of hidden nodes, in this case eight, and it is problem specific. A total of 270 data points for each of the 14 variables were provided with 136 of each of these used to train the network. The classification variable, being the Heart Disease status, was assumed to have all 67 (25%) of its test data entries missing at random and the aim was to accurately predict and classify these individuals from having prior knowledge of how the other variables affect the heart disease status. To measure the level of accuracy with which these algorithms are able to predict the expected outputs and classify individuals correctly, ROC curves are plot and the areas under these are calculated.

**Results and Discussion**

The estimates used to measure the modelling quality for the prediction problems are the Mean Square Error (MSE), Root Mean Square Error (RMSE), and Mean Absolute Error (MAE).

MSE is a measure of the square of the differences between the estimated output values and actual output values for the variable being considered. It provides a means of quantifying these differences. The smaller the MSE value, the better the prediction accuracy and vice versa. For a given dataset, the Mean Square Error (MSE) can be computed as follows:

$$MSE = \frac{1}{n}\sum_{i=1}^{N}(x_i - \hat{x}_i)^2$$

RMSE is simply the mean square error in the original predicted value units and is regarded as a quadratic scoring rule that measures the average magnitude of the error. It is obtained by calculating the square root of the variance, known as the standard deviation. It differs from the MSE by providing a reduction in the variance, hence its application in this work.

The Root Mean Square Error (RMSE) can be obtained by using the formula:

$$RMSE = \sqrt{\frac{1}{n}\sum_{i=1}^{N}(x_i - \hat{x}_i)^2}$$

MAE is a measure which takes care of possible overestimation due to the presence of outliers in the dataset, and provides a statistical measure of how far estimates or forecasts are from the actual values. It measures the average magnitude of the errors in a dataset without considering direction. The Mean Absolute Error (MAE) can be computed by using the following formula:

$$MAE = \frac{1}{n}\sum_{i=1}^{N}|x_i - \hat{x}_i|^2$$

In the equations above, N is the number of values present in the vector, {$x_i$} is the actual value being considered at that specific instance, and {$\hat{x}_i$} is the corresponding approximated value.

To measure the classification capabilities of the algorithms, the Receiver Operating Characteristic (ROC) curve is plot, and the area under this is calculated. An ROC curve is a plot which graphically illustrates the performance of a binary classifier system as there is a variation in its discrimination threshold. ROC curves are obtained by plotting the fraction of true positives out of the positives (TPR) and false positives out of the negatives (FPR) at various threshold settings. The area under the ROC curve (AUC) gives a measure of the accuracy and level of efficiency with which the algorithms are able to correctly classify unseen records.

**Experiment 1: German Credit Data**

Statistical analysis of the results obtained was performed using the t-test. The null hypothesis (H₀) states that there is no significant difference in the means of the predictions obtained by the GA, SA, PSO, RF and NS algorithms, for a population from which the sample used for analysis was obtained. The alternative hypothesis (H$_A$) on the other hand states that there is a significant difference in the means of the predictions obtained by these five algorithms.

TABLE II. STATISTICAL ANALYSIS OF THE GERMAN CREDIT DATA RESULTS

| Pairs Compared | P-Values (95% Confidence Level) |
| --- | --- |
| GA-SA | 0.19 |
| GA-PSO | 0.00 |
| GA-RF | 0.48 |
| GA-NS | 0.00 |
| SA-PSO | 0.00 |
| SA-RF | 0.45 |
| SA-NS | 0.01 |
| PSO-RF | 0.00 |
| PSO-NS | 0.13 |
| RF-NS | 0.00 |

It can be seen in the above table that there is no significant difference at a 95% confidence level in the means of the predictions between the GA and SA algorithms (p-value: 0.19), and the GA and RF algorithms (p-value: 0.48). Findings however suggest that there is a significant difference in the means of predictions between the GA algorithm and the PSO and NS algorithms (p-value: 0.00 between GA and PSO, and p-value: 0.00 between GA and NS). This therefore indicates that the null hypothesis ($H_O$), which assumes that there is no difference in the means between the GA algorithm and the RF and NS algorithms, can be rejected in favour of the alternative hypothesis ($H_A$) at a 95% confidence level.

Furthermore, it can be observed that there is a significant difference in the means of the predictions between the SA algorithm and the PSO and NS algorithms: SA and PSO (p-value: 0.00), and SA and NS (p-value: 0.01). This therefore suggests that the null hypothesis ($H_O$) can be rejected in favour of the alternative hypothesis ($H_A$) at a 95% confidence level. However, findings indicate that there is no significant difference between the means obtained by the SA and RF algorithms in the population at a confidence level of 95% with a p-value of 0.45.

Moreover, the table reveals that there is a significant difference in the means of the predictions between the PSO algorithm and the RF algorithm (p-value: 0.00). However, findings suggest that there is no significant difference between the means of the predictions obtained by the PSO and the NS algorithms with a p-value of 0.13, resulting in the null hypothesis being accepted at a 95% confidence level.

Finally, it is observed in the table that there is a significant difference in the means of the predictions between the RF and NS algorithms (p-value: 0.00), therefore, the null hypothesis ($H_O$) is rejected in favour of the alternative hypothesis ($H_A$) at a 95% confidence level.

**Experiment 2: Forest Fire Data**

Statistical analysis of the results obtained was performed using the t-test. The null hypothesis ($H_O$) states that there is no significant difference in the means of the predictions obtained by the GA, SA, PSO, RF and NS algorithms, for a population from which the sample used for analysis was obtained.

The alternative hypothesis ($H_A$) on the other hand states that there is a significant difference in the means of the predictions obtained by these five algorithms.

TABLE III. STATISTICAL ANALYSIS OF THE FOREST FIRE DATA RESULTS

| Pairs Compared | P-Values (95% Confidence Level) |
|---|---|
| GA-SA | 0.69 |
| GA-PSO | 0.35 |
| GA-RF | 0.00 |
| GA-NS | 0.00 |
| SA-PSO | 0.61 |
| SA-RF | 0.00 |
| SA-NS | 0.00 |
| PSO-RF | 0.00 |
| PSO-NS | 0.00 |
| RF-NS | 0.00 |

It can be seen from the above table that there is no significant difference at a 95% confidence level in the means of the predictions between the GA and SA algorithms (p-value: 0.69), and the GA and PSO algorithms (p-value: 0.35). However, findings suggest that there is a significant difference in the means of predictions between the GA algorithm and the RF and NS algorithms (p-value: 0.00 between GA and RF, and p-value: 0.00 between GA and NS). This therefore indicates that the null hypothesis ($H_O$), which assumes that there is no difference in the means between the GA algorithm and the RF and NS algorithms, can be rejected in favour of the alternative hypothesis ($H_A$) at a 95% confidence level.

Furthermore, it can be observed in the above table that there is a significant difference in the means of the predictions between the SA algorithm and the RF and NS algorithms: SA and RF (p-value: 0.00), and SA and NS (p-value: 0.00). However, findings indicate that there is no significant difference between the means obtained by the SA and PSO algorithms at a confidence level of 95% with a p-value of 0.61.

Moreover, the table shows that there is a significant difference in the means of the predictions between the PSO algorithm and the RF and NS algorithms (p-value: 0.00 between PSO and RF, and p-value: 0.00 between PSO and NS). This therefore suggests that the null hypothesis ($H_O$) which assumes that there is no difference in the means between the PSO algorithm and the RF and NS algorithms can be rejected in favour of the alternative hypothesis ($H_A$) at a 95% confidence level.

Finally, it can be seen in the above table that there is a significant difference in the means of the predictions between the RF and NS algorithms (p-value: 0.00), therefore, the null hypothesis ($H_O$) is rejected in favour of the alternative hypothesis ($H_A$) at a 95% confidence level.

**Experiment 3: Heart Disease Data**

Statistical analysis of the results obtained was performed using the t-test. The null hypothesis ($H_O$) states that there is no significant difference in the means of the predictions obtained by the GA, SA,

PSO, RF and NS algorithms, for a population from which the sample used for analysis was obtained. The alternative hypothesis ($H_A$) on the other hand states that there is a significant difference in the means of the predictions obtained by these five algorithms.

TABLE IV. STATISTICAL ANALYSIS OF THE FOREST FIRE DATA RESULTS

| Pairs Compared | P-Values (95% Confidence Level) |
|---|---|
| GA-SA | 0.99 |
| GA-PSO | 0.23 |
| GA-RF | 0.05 |
| GA-NS | 0.87 |
| SA-PSO | 0.22 |
| SA-RF | 0.04 |
| SA-NS | 0.87 |
| PSO-RF | 0.61 |
| PSO-NS | 0.24 |
| RF-NS | 0.02 |

The above table reveals that there is no significant difference at a 95% confidence level in the means of the predictions between the GA and SA algorithms (p-value: 0.99), the GA and PSO algorithms (p-value: 0.23), and the GA and NS algorithms (p-value: 0.87). However, findings suggest that there is a significant difference in the means of predictions between the GA and RF algorithms (p-value: 0.05). This therefore indicates that the null hypothesis ($H_O$), which assumes that there is no difference in the means between the GA algorithm and RF algorithm, can be rejected in favour of the alternative hypothesis ($H_A$) at a 95% confidence level.

Furthermore, the table shows that there is no significant difference at a 95% confidence level in the means of the predictions between the SA and PSO algorithms (p-value: 0.22), and the SA and NS algorithms (p-value: 0.87). Findings however suggest that there is a significant difference in the means of predictions between the SA and RF algorithms (p-value: 0.04). This therefore indicates that the null hypothesis ($H_O$), which assumes that there is no difference in the means between the SA algorithm and RF algorithm, can be rejected in favour of the alternative hypothesis ($H_A$) at a 95% confidence level.

Moreover, it can be seen that there is no significant difference at a 95% confidence level in the means of the predictions between the PSO and RF algorithms (p-value: 0.61), and the PSO and NS algorithms (p-value: 0.24).

Finally, the table reveals that there is a significant difference in the means of the predictions between the RF and NS algorithms (p-value: 0.02), therefore, the null hypothesis ($H_O$) is rejected in favour of the alternative hypothesis ($H_A$) at a 95% confidence level.

**Conclusion**

This research evaluates the effectiveness of using autoassociative neural networks and the Genetic Algorithm, Simulated Annealing, Particle Swarm Optimization, Random Forest, and Negative

Selection algorithms in carrying out predictions and classifications of missing data entries from datasets. The approach used in solving the prediction problems involved the approximation of the values of a continuous variable. For the classification problems, the values to be approximated and determined all emerged from categorical/discrete variables. To evaluate the effectiveness of these methods, three distinct datasets were used all with the aim of solving two main problems being; prediction and classification problems. For the prediction problem, the Forest Fire dataset was used to predict the missing values from a continuous variable being the Area affected by a fire. The other two datasets being the German Credit and Heart Disease datasets were used to test the performances of the algorithms as classifiers. For the German Credit dataset, the aim was to classify new records into the right credit status of which there were only two (two classes or categories). The Heart Disease dataset had as output the health status of an individual with two categories depicting whether or not the person with the record had a heart disease. Auto-associative neural network with five optimization algorithms being; Genetic Algorithm, Simulated Annealing, Particle Swarm Optimization, Random Forest, and Negative Selection are suggested to carry out predictions and classifications of the missing values in datasets. An auto-associative neural network is a network which is trained to predict its inputs at the output layer. An MLP neural network is used to train the auto-associative neural network using the scaled conjugate gradient method. An error function is obtained when the network is trained by calculating the square of the difference between the predicted output values from the network, and the actual output values. Since one of the components of the input vector is assumed missing, the error function is broken down so as to depend on both the known and the unknown input vector components. The optimization algorithms are then applied at this stage to find approximations of the missing data entries in the input vector with the aim of minimizing the error function.

Overall analyses of the results obtained from the experiments reveal that the Random Forest algorithm performed better than the Genetic Algorithm, Simulated Annealing algorithm, Particle Swarm Optimization algorithm, and Negative Selection algorithm. More precisely, analyses carried out on the results obtained for the prediction problem using the MSE, RMSE, and MAE values from the Forest Fire dataset revealed that the GA, SA, and PSO algorithms were observed to perform better than the RF algorithm although not statistically significant. However, when the performances of these four algorithms were compared against those of the NS algorithm for this dataset, the differences in these were significantly different as seen in Appendices A and B.

For the classification problem using the two related datasets mentioned in the previous subsection, it was observed that the RF algorithm performed better than the GA, SA, PSO, and NS algorithms. The most significant differences in performance were observed when the AUC values for the RF algorithm were compared to those obtained by using the NS algorithm which produced the lowest AUC values on both datasets. The reason for the Random Forest algorithm doing very well compared to the other algorithms could be attributed to the fact that it is an ensemble approach or method. Ensemble methods use multiple models to obtain better predictive performance than could be obtained from any of the constituent models or from using just one method. An ensemble is a supervised learning algorithm due to the fact that it can be trained and then used to make predictions. Another advantage of the Random Forest algorithm over the other algorithms is the fact that during training, it is not sensitive to outliers in the training data, and there is no guarantee of such robustness to outliers from the GA, SA, PSO, and NS algorithms. Furthermore, with the Random Forest algorithm, there is no risk of over-fitting the data and setting the parameters is easy considering that there are very few parameters, one exactly in this research being the number of trees, which has no effect, regardless of what the value is on its predictive and classification capability. GA, SA, PSO, and NS all have several parameters which if not properly set, could lead to poor solutions being obtained, thus setting these parameters is quite vital. The most noticeable advantage of the Random Forest algorithm over the other algorithms is the fact that it is very fast in converging to a guaranteed optimal global solution.

In this paper, we used neural networks as a learning method to train the datasets. Though the choice for neural network was as a result of its merits compared to other machine learning techniques and the fact that it has produced good solutions, it is worthwhile trying to investigate and construct the models using other machine learning techniques. One of these which could be used to solve the same problem could be to train the model using Decision Trees or Support Vector Machines instead of neural networks. Other optimization algorithms could be investigated such as ant colony optimization, Bees algorithm or Coocoo search.